\pdfoutput=1

\documentclass[11pt]{article}

\usepackage[]{acl}

\usepackage{times}
\usepackage{latexsym}

\usepackage[T1]{fontenc}

\usepackage[utf8]{inputenc}

\usepackage{microtype}

\usepackage{amsmath}
\usepackage{amssymb}
\usepackage{amsfonts}
\usepackage{bbm}
\usepackage{xfrac}
\usepackage{multirow}
\usepackage{graphicx}
\usepackage{subcaption}

\newcommand{\ssz}{\scriptsize}
\newcommand{\ftn}{\tiny}

\title{Gender Biases and Where to Find Them:\\ 
Exploring Gender Bias in Pre-Trained Transformer-based Language Models Using Movement Pruning}

\author{Przemyslaw Joniak \\
    The University of Tokyo \\
    \texttt{joniak@g.ecc.u-tokyo.ac.jp} \\\And
    Akiko Aizawa \\
    National Institute of Informatics \\
    \texttt{aizawa@nii.ac.jp} \\}

\begin{document}
\maketitle
\begin{abstract}
Language model debiasing has emerged as an important field of study in the NLP community. Numerous debiasing techniques were proposed, but bias ablation remains an unaddressed issue. We demonstrate a novel framework for inspecting bias in pre-trained transformer-based language models via movement pruning. Given a model and a debiasing objective, our framework finds a subset of the model containing less bias than the original model. We implement our framework by pruning the model while fine-tuning it on the debiasing objective. Optimized are only the pruning scores --- parameters coupled with the model's weights that act as gates. We experiment with pruning attention heads, an important building block of transformers: we prune square blocks, as well as establish a new way of pruning the entire heads. Lastly, we demonstrate the usage of our framework using gender bias, and based on our findings, we propose an improvement to an existing debiasing method. Additionally, we re-discover a bias-performance trade-off: the better the model performs, the more bias it contains.
\end{abstract}

\section{Introduction}
Where in language models (LM) is bias stored? Can a~neural architecture itself impose a bias? There is no consensus on this matter. \citet{kaneko-bollegala-2021-debiasing} suggest that gender bias resides on every layer of transformer-based LMs. However, this is somehow vague — transformer layers can be further decomposed into building blocks, namely attention heads, and these also can be further broken down into matrices.
On the other hand, the findings of \citet{voita-etal-2019-analyzing} show that some attention heads within layers specialize in particular tasks, such as syntactic and positional dependencies. 
This gives us an intuition that some heads, or their parts, may specialize in learning biases as well. 
Being able to analyze bias in language models on a more granular level, would bring us a better understanding of the models and the phenomenon of bias. With knowledge of where the bias is stored, we could design debiasing techniques that target particular parts of the model,
making the debiasing more accurate and efficient.

We demonstrate a novel framework that utilizes movement pruning \cite{Sanh2020MovementPA} to inspect biases in language models.
Movement pruning was originally used to compress neural models and make its inference faster. 
We introduce a modification of movement pruning that enables us to choose a low-bias subset of a given model, or equivalently, find these model's weights whose removal leads to convergence of an arbitrary debiasing objective.
Specifically, we freeze neural weights of the model and optimize only the so-called pruning scores that are coupled with the weights and act as gates. 
This way, we can inspect which building blocks of the transformers, i.e. attention heads, might induce bias.
If a head is pruned and the debiasing objective converges, then we hypothesize that the head must have contained bias.
We demonstrate the utility of our framework using \citet{kaneko-bollegala-2021-debiasing}'s method of removing gender bias.

Biases have been extensively studied and numerous debiasing methods were proposed. In fact, according to \citet{Stanczak2021ASO}, the ACL Anthology saw an exponential growth of bias-related publications in the past decade -- and it only counts gender bias alone. 
Nonetheless, the vast majority of these works address problems of bias detection or mitigation only.
To our best knowledge, we are the first to conduct bias ablation in LMs.
We:
(1)~demonstrate an original framework to inspect biases in LMs. Its novelty is a mixture of movement pruning, weight freezing and debiasing;
(2)~study the presence of gender bias in a BERT model;
(3)~propose an improvement to an existing debiasing method, and
(4)~release our code\footnote{\url{https://github.com/kainoj/pruning-bias}}.

\begin{table*}[]
\centering
\setlength{\tabcolsep}{3pt}
\setlength{\arrayrulewidth}{0.001pt}
\renewcommand{\arraystretch}{1.03}
\footnotesize
\begin{tabular}{crr|rrrr|rrrrrrrrrr|r}
\multicolumn{1}{c}{\ssz Block}      & \ssz Layer & \ssz Mode     & 
\ssz SEAT6 & \ssz SEAT7 & \ssz SEAT8 & \ssz SS & \tiny COLA & \tiny SST2 & \tiny MRPC & \tiny STSB & \tiny QQP & \tiny MNLI & \tiny QNLI & \tiny RTE & \tiny WNLI & \ssz GLUE &\ssz $\#P$\\ \hline
\multirow{4}{*}{32x32}         & all   & token    & 0.91  & 0.95  & 0.92  & 51.9	& 0.0   & 87.2 & 73.6 & 46.7 & 86.8 & 77.6 & 83.2 &	55.2 &	49.3 &	62.2 & 0 \\
                               &       & sentence & 0.67  & -0.40 & -0.23 & \textbf{49.5}	& 2.7   & 87.8 & 75.4 & 63.2 & 86.6 & 76.2 & 83.5 &	54.2 &	54.9 &	64.9 & 0 \\
                               & last  & token    & 1.39  & 0.57  & 0.18  & 52.6	& 15.5	& 90.1 & 75.8 & 82.2 & 86.8 & 79.5 & 85.6 &	57.0 &	42.3 &	68.3 & 0 \\
                               &       & sentence & 0.85  & 0.64  & 0.67  & 51.9	& 9.0	& 89.1 & 75.1 & 77.4 & 87.1 & 79.3 & 86.1 &	56.7 &	39.4 &	66.6 & 0 \\ \hline
\multirow{4}{*}{64x64}         & all   & token    & 0.43  & 0.22  & \textbf{0.01}  & 53.4	& 4.7	& 86.5 & 74.7 & 76.9 & 86.4 & 77.3 & 83.6 &	54.5 &	43.7 &	65.4 & 1 \\
                               &       & sentence & 0.28  & 0.56  & -0.06 & 49.3	& 5.9	& 86.6 & 73.9 & 79.1 & 86.0 & 77.2 & 83.0 &	54.5 &	47.9 &	66.0 & 1 \\
                               & last  & token    & 0.67  & -0.31 & -0.36 & 51.9	& 0.0	& 86.4 & 76.0 & 80.2 & 86.4 & 78.1 & 83.7 &	52.7 &	42.3 &	65.1 & 0 \\
                               &       & sentence & 0.72  & 0.57  & 0.03  & 56.0	& 4.6	& 89.2 & 75.7 & 84.0 & 87.0 & 79.4 & 85.3 &	52.3 &	39.4 &	66.3 & 0 \\ \hline
\multirow{4}{*}{128x128}       & all   & token    & 0.84  & 0.47  & 0.17  & 50.9	& 3.3	& 87.2 & 74.2 & 69.9 & 86.4 & 77.7 & 83.8 &	53.1 &	50.7 &	65.1 & 6 \\
                               &       & sentence & 0.55  & 0.17 & 0.22  & 54.2	& 6.6	& 85.4 & 75.0 & 79.3 & 85.7 & 76.9 & 83.0 &	56.0 &	42.3 &	65.6 & 8 \\
                               & last  & token    & 0.65  & 0.17 & -0.13 & 49.1	& 0.3	& 85.6 & 76.8 & 44.3 & 86.3 & 76.7 & 82.9 &	52.3 &	\textbf{56.3} &	62.4 & 2 \\
                               &       & sentence & 0.10  & 0.35  & -0.22 & \textbf{49.5}	& 0.0	& 84.4 & 73.8 & 75.7 & 85.6 & 77.2 & 82.7 &	43.7 &	52.1 &	63.9 & 2 \\ \hline
\multirow{4}{*}{\shortstack{$64\times 768$ \\ \tiny{(entire head)}}}
                               & all   & token    & 0.75  & 0.49  & 0.29  & 57.2	& \textbf{38.8}	& \textbf{91.4} & 78.3 & 86.3 & 88.5 & \textbf{82.9} & 88.6 & 57.0 & \textbf{56.3} &	\textbf{74.3} & 61 \\
                               &       & sentence & 0.48  & -0.17 & 0.02  & 56.0	& 26.9	& 90.6 & 79.2 & \textbf{86.5} & 88.4 & 83.4 & 88.9 & 57.4 &	40.8 &	71.3 & 66 \\
                               & last  & token    & 0.62  & -0.17 & -0.27 & 58.5	& 44.6	& \textbf{91.4} & \textbf{78.5} & 81.4 & \textbf{88.6} & 82.0 & \textbf{88.9} & \textbf{58.1} & 52.1 &	74.0 & 58 \\
                               &       & sentence & \textbf{0.09}  &\textbf{ 0.05}  & 0.34  & 58.7	& 36.7	& 91.3 & 76.9 & 84.7 & 87.8 & 81.5 & 87.9 &	50.9 &	43.7 &	71.3 & \textbf{93} \\ \hline
       -          & \multicolumn{2}{c|}{original} & 1.04  & 0.22  & 0.63  & 62.8    & 58.6  & 92.8 & 87.2 & 88.5 & 89.4 & 85.1 & 91.5 & 64.3 &  56.3 &  79.3 & -
\end{tabular}
\caption{\label{tab:prunedbias}
Bias in fine-pruned models for various block sizes, evaluated using \textit{SEAT} and \textit{stereotype score} (SS). 
Ideally, bias-free model has a SEAT of 0 and SS of 50.
GLUE evaluated using only these weights in a model that were not pruned.
$\#P$ indicates number of heads that were entirely pruned.
Best fine-pruning results are in \textbf{bold}.
}
\end{table*}

\section{Background}

\subsection{Language Model Debiasing}
\label{sec:debiasing}
Numerous paradigms for language model debiasing were proposed, including 
feature extraction-based \cite{Pryzant2020AutomaticallyNS}, 
data augmentations \cite{zhao-etal-2019-gender, Lu2020GenderBI, dinan-etal-2020-queens}, 
or paraphrasing \cite{ma-etal-2020-powertransformer}.
They all require an extra endeavor, such as feature engineering, re-training, or building an auxiliary model.

We choose an algorithm by \citet{kaneko-bollegala-2021-debiasing} for removing gendered stereotypical associations. It is competitive, as it can be applied to many transformer-based models, and requires minimal data annotations.
The algorithm enforces embeddings of predefined gendered words (e.g. \textit{man}, \textit{woman}) to be orthogonal to their stereotyped equivalents (e.g. \textit{doctor}, \textit{nurse}) via fine-tuning.
The loss function is a squared dot product of these embedding plus a regularizer between the original and the debiased model. 
The former encourages orthogonality and the latter helps to preserve syntactic information.

The authors proposed six debiasing modes: \texttt{all-token}, \texttt{all-sentence}, \texttt{first-} \texttt{token}, \texttt{first-sentence}, \texttt{last-token}, and \texttt{last-sentence}, depending on source of the embeddings (first, last or all layers of a transformer-based model) and target of the loss (target token or all tokens in a sentence). In this work, we omit the \texttt{first-*} modes, as they were shown to have an insignificant debiasing effect.

\subsection{Block Movement Pruning}
Pruning is a general term used when disabling or removing some weights from a neural network. 
It can lead to a higher sparsity, making a model faster and smaller while retaining its original performance.
Movement pruning, introduced by \citealp{Sanh2020MovementPA} discards a weight when it \textit{moves} towards zero.
\citealp{lagunas-etal-2021-block} proposed pruning entire blocks of weights: with every weight matrix
$W\in\mathbb{R}^{M\times N}$,
a~score matrix $S\in\mathbb{R}^{\sfrac{M}{M'}\times\sfrac{N}{N'}}$ is associated, where $(M', N')$ is a pruning block size.
On the forward pass, $W$ is substituted with its masked version, $W' \in\mathbb{R}^{M\times N}$:
\begin{align*} 
    W' &= W \odot \mathbf{M}(S) \\
    \mathbf{M}_{i, j} &= \mathbbm{1}
                        \Big(
                        \sigma\big(
                            S_{
                                \lceil \sfrac{i}{M'} \rceil,
                                \lceil \sfrac{j}{N'} \rceil
                              }
                        \big)
                        > \tau
                        \Big),
\end{align*}
where $\odot$ stands for element-wise product, $\sigma$ is the Sigmoid function, $\tau$ is a threshold and $\mathbbm{1}$ denotes the indicator function.
On the backward pass, both $W$ and $S$ are updated.
To preserve the performance of the original model, \citet{lagunas-etal-2021-block} suggest using a teacher model as in the model distillation technique \cite{Sanh2019DistilBERTAD}.

We decided to utilize movement pruning because of the mechanism of the scores $S$.
The scores can be optimized independently of weights, and thus we can freeze the weights.
This would be impossible with e.g. magnitude pruning \cite{Han2015LearningBW} which directly operates on weights values (magnitudes).

\section{Exploring Gender Bias Using Movement Pruning}
We focus on gender bias defined as stereotypical associations between male and female entities. Our study is limited to the English language and binary gender only.

We attempt to answer the following questions: in transformer-based pre-trained language models, 
can we identify particular layers or neighboring regions that are in charge of biases?
To verify this, we propose a simple and, to our best knowledge, novel framework based on debiasing and attention head block movement pruning.
Given a pre-trained model and a fine-tuning objective, we find which attention blocks can be disabled, so the model performs well on the task. 
We prune the model while fine-tuning it on a debiasing objective, such as the one described in $\S$\ref{sec:debiasing}. We optimize solely the pruning scores $S$ and the weights $W$ of the original model remain untouched (they are \emph{frozen}). 

We target the building blocks of transformer-based models, attention heads \cite{Vaswani2017AttentionIA}. 
Each head consists of four learnable matrices, and we prune all of them. In $\S$\ref{sec:exp}, we test two strategies: pruning square blocks of the matrices and pruning entire attention heads.

To evaluate bias, we utilize Sentence Encoder Association Test
(SEAT, \citet{may-etal-2019-measuring} and StereoSet Stereotype Score (SS, \citet{nadeem-etal-2021-stereoset} evaluated on the gender domain.
To measure model performance, we utilize GLUE \cite{wang-etal-2018-glue}, a standard NLP benchmark.

\subsection{Experiments}\label{sec:exp}
In all experiments, we use the BERT-base model \cite{devlin-etal-2019-bert}.
See Appendix for used datasets and detailed hyperparameters.

\paragraph{Square Block Pruning.}
\begin{figure}[]
    \centering
    \includegraphics[width=1\linewidth]{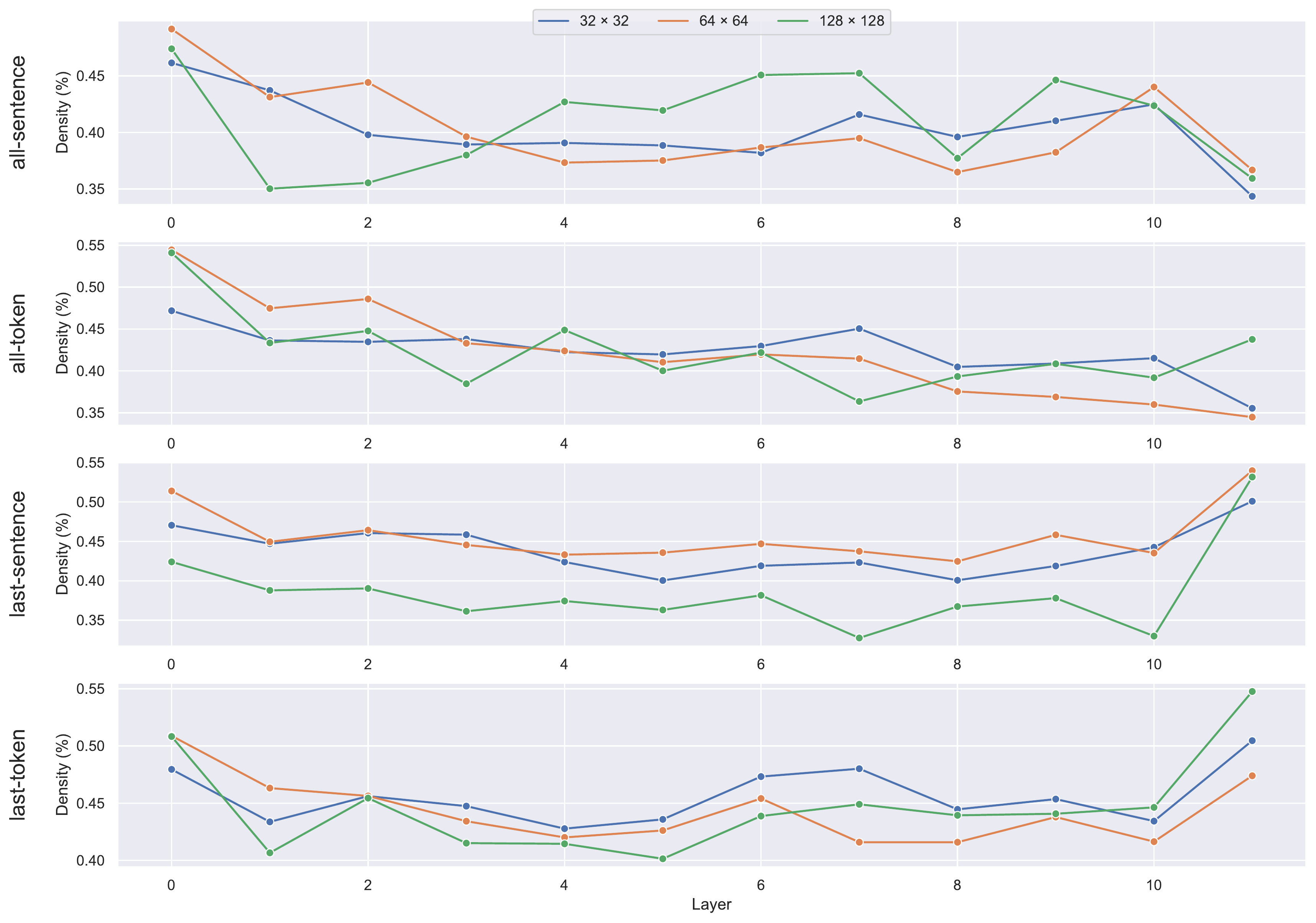}
    \caption{\label{fig:avg-densities}
     Per-layer densities of fine-pruned models using different debiasing modes, for multiple square block sizes.
     Density is computed as a percentage of non-zero elements within a layer.
    }
\end{figure}
\citet{lagunas-etal-2021-block} showed that square block pruning in attention head matrices leads to the removal of whole attention heads. Although our objective differs from theirs, we attempt to reproduce this behavior.
To find the best square block size $(B, B)$, we experiment with  $B = 32, 64, 128$.
See Tab. \ref{tab:prunedbias}.
We also tried with $B = 256, 384$, and  $768$, but we discarded these values as we faced issues with convergence. 
Choosing a suitable block size is a main limitation of our work.

\paragraph{Attention Head Pruning.}\label{section:att-head-prune}
To remove entire attention heads, we cannot prune all head matrices at once -- see Appendix for a detailed explanation.
Instead, we prune $64 \times 768$ blocks (size of the attention head in the BERT-base) of the \textit{values} matrices solely. See the last row group of Tab. \ref{tab:prunedbias} for the results.
%
\begin{figure}[]
    \centering
    \includegraphics[width=\linewidth]{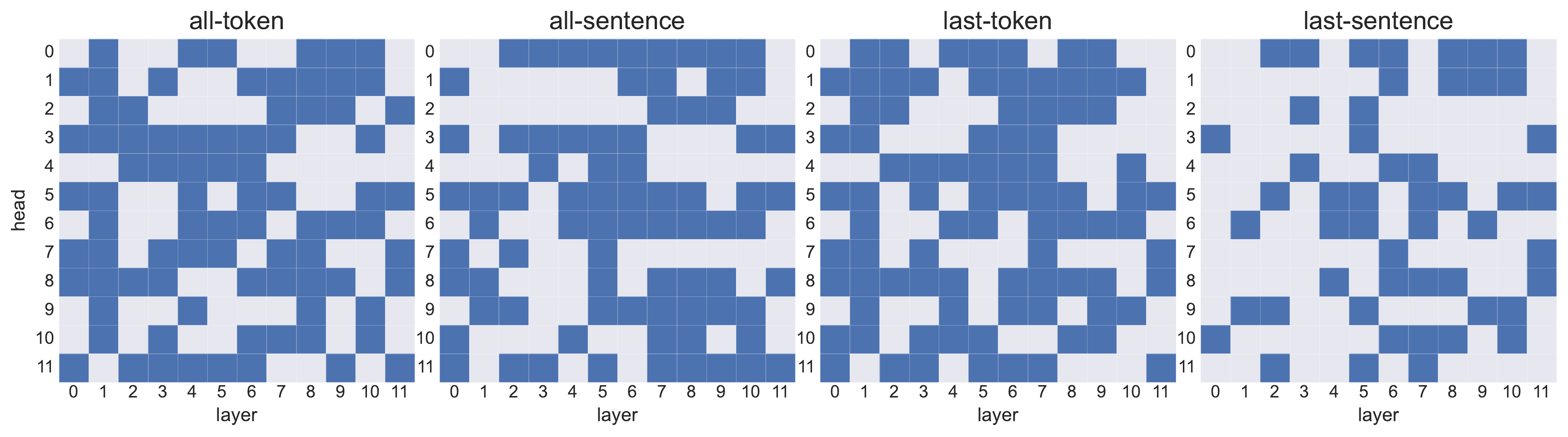}
    \caption{Pruning entire heads: which heads remained (blue) and which heads were pruned (gray)?
    }
    \label{fig:prune-heatmap}
\end{figure}
%

\subsection{Discussion}\label{sec:discussion}

\paragraph{Square Block Pruning Does Not Remove Entire Heads}
\citealp{lagunas-etal-2021-block} found that pruning square block removes entire heads.
However, we failed to observe this phenomenon in the debiasing setting--see last column of Tab~\ref{tab:prunedbias}.
We are able to prune at most $8$ heads, only for relatively large block sizes, $128\times 128$.
We hypothesize that the reason is the weight freezing of the pre-trained model.
To verify this, we repeat the experiment with $32\times 32$ block size, but we do not freeze the weights.
Bias did not change significantly, but no attention heads were fully pruned (Tab. \ref{tab:no-freeze}). This suggests that bias may not be encoded in particular heads, but rather is distributed over multiple heads.
\begin{table}[]
\centering
\setlength{\tabcolsep}{1pt}
\footnotesize
\begin{tabular*}{\linewidth}{l@{\extracolsep{\fill}}rrrrrrr}
     &       & SEAT6\tiny{$(\Delta)$} & SEAT7\tiny{$(\Delta)$} & SEAT8\tiny{$(\Delta)$} & GLUE\tiny{$(\Delta)$}  & $\#P$  \\ \hline 
all  & token & 1.25 \tiny{(+0.3)} & 0.54 \tiny{(-0.4)} & 0.58 \tiny{(-0.3)} & 74.0 \tiny{(+12)} & 0 \\
     & sent. & 1.10 \tiny{(+0.4)} & 0.48 \tiny{(+0.1)} & 0.18 \tiny{(+0.0)} & 71.5 \tiny{(+7)} & 0 \\
last & token & 1.31 \tiny{(-0.1)} & 0.43 \tiny{(-0.1)} & 0.45 \tiny{(+0.3)} & 74.4 \tiny{(+6)} & 0 \\
     & sent. & 1.24 \tiny{(+0.4)} & 0.82 \tiny{(+0.2)} & 0.72 \tiny{(+0.0)} & 72.2 \tiny{(+6)} & 0 
\end{tabular*}
\caption{\label{tab:no-freeze}
SEAT, GLUE, and number of fully pruned attention heads ($\#P$) for the
$32\times 32$ block pruning when allowing the weight of the model to change.
$\Delta$~refers to a relative change to results in Tab. \ref{tab:prunedbias}, that is when the original weights are frozen.
See Appx. Tab.~\ref{tab:all-no-freeze} for the full GLUE breakdown.
}
\end{table}

\begin{figure}[h!]
    \centering
    \includegraphics[width=0.99\linewidth]{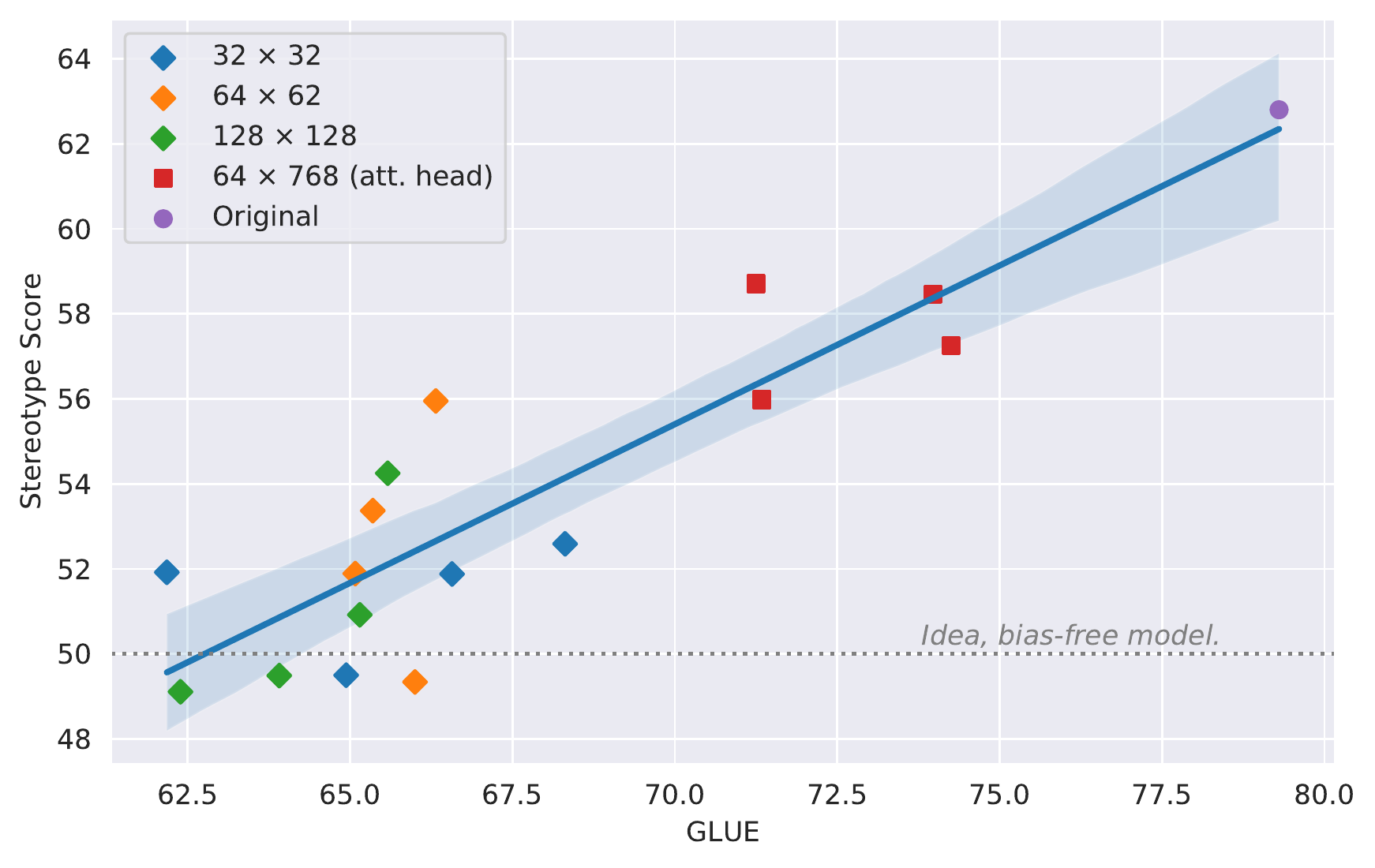}
    \caption{%
    Performance-bias trade-off for various models.
    The better a model performs, the more bias it has.
    }
    \label{fig:tradeoff-glue-vs-ss-tradeoff}
\end{figure}
\paragraph{Performance-Bias Trade-off}%
We observe that there is a negative correlation between model performance and its bias (Fig. \ref{fig:tradeoff-glue-vs-ss-tradeoff}).
Models that contain no bias, i.e. with SS close to 50, perform poorly. The model with the best GLUE contains the most bias.
This phenomenon might be an inherent weakness of the debiasing algorithm.
To alleviate the issue, it might be necessary to improve the algorithm, work on a better one, or focus on debiasing data. 
It would be also interesting to try optimizing the debiasing and the downstream task objective simultaneously. However, this is out of the scope of our study and we leave it for future work.

\paragraph{The Failure of the CoLA GLUE Task}\label{sec:cola-fail}
Our models perform  poorly on the Corpus of Linguistic Acceptability task (CoLA, \citet{warstadt-etal-2019-neural}). Most of them have scores close to zero, meaning i.e they take a random, uninformed guess. The reason might lay in the complexity of the task. CoLA remains the most challenging task out of the whole GLUE suite as it requires deep syntactic and grammatical knowledge. It has been suggested that language models do not excel at grammatical reasoning \cite{Sugawara2020AssessingTB},
and it~might be that perturbations such as the absence of the weights (pruning) break already weak grammatical abilities.
The results in Tab.~\ref{tab:no-freeze} support this hypothesis. Compared to the `frozen' setting, CoLA scores are significantly higher, whereas the other tasks see just a slight increase (Tab.~\ref{tab:no-freeze}).

\section{Debiasing Early Intermediate Layers Is Competitive}
\citet{kaneko-bollegala-2021-debiasing} proposed three heuristics: debiasing the first, last, and all layers.
However, the number of layer subsets that can be debiased is much larger.
Trying all subsets to find the best one is prohibitively expensive.
With our framework, we are able to find a better subset with a low computational cost.

We observed that: (1) square block pruning does not significantly affect the first and last layer: densities of these layers are usually higher than the other layers' (Fig.~\ref{fig:avg-densities});  (2) attention head pruning mostly affects intermediate layers (Fig.~\ref{fig:prune-heatmap}).
Based on the above, we propose to debias intermediate layers. 
Specifically, we take the embeddings from layers index 1 to 4 inclusive, and we run the debiasing algorithm described in $\S$\ref{sec:debiasing}.
We do not include layer 0 because it generally yields high densities (ref. Fig.\ref{fig:avg-densities}), and layer 5, as it contains the most number of heads that were not pruned in every experiment (ref. Fig.~\ref{fig:prune-heatmap}).
We end up with two more modes, \texttt{intermediate-token} and \texttt{intermediate-sentence}. We present results for our, as well as the other modes in Tab.~\ref{tab:interm-debs} (note that the results may differ from \citet{kaneko-bollegala-2021-debiasing}'s due to random seed choice). Debiasing the intermediate layers is competitive to debiasing \texttt{all} and \texttt{last} layers.
The SS of the \texttt{intermediate-} modes is lower that the SS of corresponding
\texttt{all} and \texttt{last} modes. The SS of \texttt{intermediate-sentence} gets close to the perfect score of 50.
\begin{table}[]
\centering
\setlength{\tabcolsep}{2pt}
\footnotesize
\begin{tabular}{rrrrrrr}
\small Layer & \small Mode   & \small SEAT6 & \small SEAT7  & \small SEAT8  &\small  $\;\;\;$ SS  & \small GLUE \\ \hline
all         & token          & 1.02  & 0.22   & 0.63   & 61.5          & 78.7            \\
            & sent.          & 0.98  & -0.34  & -0.29  & 56.9          & 75.2            \\
last        & token          & 0.98  & \textbf{0.12}   & 0.79   & 60.9          & 78.6            \\
            & sent.          & \textbf{0.39}  & -0.89  & \textbf{-0.11}  & 61.6          & \textbf{78.7}   \\ 
interm.     & token          & 1.03  & 0.33   & 0.84   & 58.5          & 77.7            \\
            & sent.          & 0.83  & 0.49   & 0.92   & \textbf{53.5} & 74.7            \\ \hline
\multicolumn{2}{c}{original} & 1.04  & 0.18   & 0.81   & 62.8  & 79.3 
\end{tabular}
\caption{\label{tab:interm-debs}
Debiasing-only results for various modes, including our original \texttt{intermediate} mode (no pruning involved).
}
\end{table}

\section{Conclusion}
We demonstrate a novel framework to inspect sources of biases in a pre-trained transformer-based language model. 
Given a model and a debiasing objective, the framework utilizes movement pruning to find a subset that contains less bias than the original model. 
We present usage of our framework using gender bias, and we found that the bias is mostly encoded in intermediate layers of BERT. 
Based on these findings, we propose two new debiasing modes that reduce more bias than existing modes. 
Bias is evaluated using SEAT and Stereotype Score metric.
Lastly, we explore a performance-bias trade-off: the better the model performs on a task, the more gender bias it has.

We hope that in the future our framework will find more applications, not only limited to gender bias.

\bibliography{anthology,custom}

\begin{thebibliography}{25}
\expandafter\ifx\csname natexlab\endcsname\relax\def\natexlab#1{#1}\fi

\bibitem[{Caliskan et~al.(2017)Caliskan, Bryson, and
  Narayanan}]{Caliskan2017SemanticsDA}
Aylin Caliskan, Joanna~J. Bryson, and Arvind Narayanan. 2017.
\newblock Semantics derived automatically from language corpora contain
  human-like biases.
\newblock \emph{Science}, 356:183 -- 186.

\bibitem[{Devlin et~al.(2019)Devlin, Chang, Lee, and
  Toutanova}]{devlin-etal-2019-bert}
Jacob Devlin, Ming-Wei Chang, Kenton Lee, and Kristina Toutanova. 2019.
\newblock \href {https://doi.org/10.18653/v1/N19-1423} {{BERT}: Pre-training of
  deep bidirectional transformers for language understanding}.
\newblock In \emph{Proceedings of the 2019 Conference of the North {A}merican
  Chapter of the Association for Computational Linguistics: Human Language
  Technologies, Volume 1 (Long and Short Papers)}, pages 4171--4186,
  Minneapolis, Minnesota. Association for Computational Linguistics.

\bibitem[{Dinan et~al.(2020)Dinan, Fan, Williams, Urbanek, Kiela, and
  Weston}]{dinan-etal-2020-queens}
Emily Dinan, Angela Fan, Adina Williams, Jack Urbanek, Douwe Kiela, and Jason
  Weston. 2020.
\newblock \href {https://doi.org/10.18653/v1/2020.emnlp-main.656} {Queens are
  powerful too: Mitigating gender bias in dialogue generation}.
\newblock In \emph{Proceedings of the 2020 Conference on Empirical Methods in
  Natural Language Processing (EMNLP)}, pages 8173--8188, Online. Association
  for Computational Linguistics.

\bibitem[{Han et~al.(2015)Han, Pool, Tran, and Dally}]{Han2015LearningBW}
Song Han, Jeff Pool, John Tran, and William~J. Dally. 2015.
\newblock Learning both weights and connections for efficient neural network.
\newblock \emph{ArXiv}, abs/1506.02626.

\bibitem[{Kaneko and Bollegala(2019)}]{kaneko-bollegala-2019-gender}
Masahiro Kaneko and Danushka Bollegala. 2019.
\newblock \href {https://doi.org/10.18653/v1/P19-1160} {Gender-preserving
  debiasing for pre-trained word embeddings}.
\newblock In \emph{Proceedings of the 57th Annual Meeting of the Association
  for Computational Linguistics}, pages 1641--1650, Florence, Italy.
  Association for Computational Linguistics.

\bibitem[{Kaneko and Bollegala(2021)}]{kaneko-bollegala-2021-debiasing}
Masahiro Kaneko and Danushka Bollegala. 2021.
\newblock \href {https://doi.org/10.18653/v1/2021.eacl-main.107} {Debiasing
  pre-trained contextualised embeddings}.
\newblock In \emph{Proceedings of the 16th Conference of the European Chapter
  of the Association for Computational Linguistics: Main Volume}, pages
  1256--1266, Online. Association for Computational Linguistics.

\bibitem[{Komisyonu(2020)}]{komisyonu2020union}
Avrupa Komisyonu. 2020.
\newblock A union of equality: Gender equality strategy 2020-2025.
\newblock \emph{2020b), https://eurlex.europa.eu/legal-content/EN/TXT}.

\bibitem[{Lagunas et~al.(2021)Lagunas, Charlaix, Sanh, and
  Rush}]{lagunas-etal-2021-block}
Fran{\c{c}}ois Lagunas, Ella Charlaix, Victor Sanh, and Alexander Rush. 2021.
\newblock \href {https://doi.org/10.18653/v1/2021.emnlp-main.829} {Block
  pruning for faster transformers}.
\newblock In \emph{Proceedings of the 2021 Conference on Empirical Methods in
  Natural Language Processing}, pages 10619--10629, Online and Punta Cana,
  Dominican Republic. Association for Computational Linguistics.

\bibitem[{Lu et~al.(2020)Lu, Mardziel, Wu, Amancharla, and
  Datta}]{Lu2020GenderBI}
Kaiji Lu, Piotr Mardziel, Fangjing Wu, Preetam Amancharla, and Anupam Datta.
  2020.
\newblock Gender bias in neural natural language processing.
\newblock \emph{ArXiv}, abs/1807.11714.

\bibitem[{Ma et~al.(2020)Ma, Sap, Rashkin, and
  Choi}]{ma-etal-2020-powertransformer}
Xinyao Ma, Maarten Sap, Hannah Rashkin, and Yejin Choi. 2020.
\newblock \href {https://doi.org/10.18653/v1/2020.emnlp-main.602}
  {{P}ower{T}ransformer: Unsupervised controllable revision for biased language
  correction}.
\newblock In \emph{Proceedings of the 2020 Conference on Empirical Methods in
  Natural Language Processing (EMNLP)}, pages 7426--7441, Online. Association
  for Computational Linguistics.

\bibitem[{May et~al.(2019)May, Wang, Bordia, Bowman, and
  Rudinger}]{may-etal-2019-measuring}
Chandler May, Alex Wang, Shikha Bordia, Samuel~R. Bowman, and Rachel Rudinger.
  2019.
\newblock \href {https://doi.org/10.18653/v1/N19-1063} {On measuring social
  biases in sentence encoders}.
\newblock In \emph{Proceedings of the 2019 Conference of the North {A}merican
  Chapter of the Association for Computational Linguistics: Human Language
  Technologies, Volume 1 (Long and Short Papers)}, pages 622--628, Minneapolis,
  Minnesota. Association for Computational Linguistics.

\bibitem[{Micikevicius et~al.(2018)Micikevicius, Narang, Alben, Diamos, Elsen,
  Garc{\'i}a, Ginsburg, Houston, Kuchaiev, Venkatesh, and
  Wu}]{Micikevicius2018MixedPT}
Paulius Micikevicius, Sharan Narang, Jonah Alben, Gregory~Frederick Diamos,
  Erich Elsen, David Garc{\'i}a, Boris Ginsburg, Michael Houston, Oleksii
  Kuchaiev, Ganesh Venkatesh, and Hao Wu. 2018.
\newblock Mixed precision training.
\newblock \emph{ArXiv}, abs/1710.03740.

\bibitem[{Nadeem et~al.(2021)Nadeem, Bethke, and
  Reddy}]{nadeem-etal-2021-stereoset}
Moin Nadeem, Anna Bethke, and Siva Reddy. 2021.
\newblock \href {https://doi.org/10.18653/v1/2021.acl-long.416} {{S}tereo{S}et:
  Measuring stereotypical bias in pretrained language models}.
\newblock In \emph{Proceedings of the 59th Annual Meeting of the Association
  for Computational Linguistics and the 11th International Joint Conference on
  Natural Language Processing (Volume 1: Long Papers)}, pages 5356--5371,
  Online. Association for Computational Linguistics.

\bibitem[{Pryzant et~al.(2020)Pryzant, Martinez, Dass, Kurohashi, Jurafsky, and
  Yang}]{Pryzant2020AutomaticallyNS}
Reid Pryzant, Richard~Diehl Martinez, Nathan Dass, Sadao Kurohashi, Dan
  Jurafsky, and Diyi Yang. 2020.
\newblock Automatically neutralizing subjective bias in text.
\newblock In \emph{AAAI}.

\bibitem[{Sanh et~al.(2019)Sanh, Debut, Chaumond, and
  Wolf}]{Sanh2019DistilBERTAD}
Victor Sanh, Lysandre Debut, Julien Chaumond, and Thomas Wolf. 2019.
\newblock Distilbert, a distilled version of bert: smaller, faster, cheaper and
  lighter.
\newblock \emph{ArXiv}, abs/1910.01108.

\bibitem[{Sanh et~al.(2020)Sanh, Wolf, and Rush}]{Sanh2020MovementPA}
Victor Sanh, Thomas Wolf, and Alexander~M. Rush. 2020.
\newblock Movement pruning: Adaptive sparsity by fine-tuning.
\newblock \emph{ArXiv}, abs/2005.07683.

\bibitem[{Stanczak and Augenstein(2021)}]{Stanczak2021ASO}
Karolina Stanczak and Isabelle Augenstein. 2021.
\newblock A survey on gender bias in natural language processing.
\newblock \emph{ArXiv}, abs/2112.14168.

\bibitem[{Sugawara et~al.(2020)Sugawara, Stenetorp, Inui, and
  Aizawa}]{Sugawara2020AssessingTB}
Saku Sugawara, Pontus Stenetorp, Kentaro Inui, and Akiko Aizawa. 2020.
\newblock Assessing the benchmarking capacity of machine reading comprehension
  datasets.
\newblock \emph{ArXiv}, abs/1911.09241.

\bibitem[{Vaswani et~al.(2017)Vaswani, Shazeer, Parmar, Uszkoreit, Jones,
  Gomez, Kaiser, and Polosukhin}]{Vaswani2017AttentionIA}
Ashish Vaswani, Noam~M. Shazeer, Niki Parmar, Jakob Uszkoreit, Llion Jones,
  Aidan~N. Gomez, Lukasz Kaiser, and Illia Polosukhin. 2017.
\newblock Attention is all you need.
\newblock \emph{ArXiv}, abs/1706.03762.

\bibitem[{Voita et~al.(2019)Voita, Talbot, Moiseev, Sennrich, and
  Titov}]{voita-etal-2019-analyzing}
Elena Voita, David Talbot, Fedor Moiseev, Rico Sennrich, and Ivan Titov. 2019.
\newblock \href {https://doi.org/10.18653/v1/P19-1580} {Analyzing multi-head
  self-attention: Specialized heads do the heavy lifting, the rest can be
  pruned}.
\newblock In \emph{Proceedings of the 57th Annual Meeting of the Association
  for Computational Linguistics}, pages 5797--5808, Florence, Italy.
  Association for Computational Linguistics.

\bibitem[{Wang et~al.(2018)Wang, Singh, Michael, Hill, Levy, and
  Bowman}]{wang-etal-2018-glue}
Alex Wang, Amanpreet Singh, Julian Michael, Felix Hill, Omer Levy, and Samuel
  Bowman. 2018.
\newblock \href {https://doi.org/10.18653/v1/W18-5446} {{GLUE}: A multi-task
  benchmark and analysis platform for natural language understanding}.
\newblock In \emph{Proceedings of the 2018 {EMNLP} Workshop {B}lackbox{NLP}:
  Analyzing and Interpreting Neural Networks for {NLP}}, pages 353--355,
  Brussels, Belgium. Association for Computational Linguistics.

\bibitem[{Warstadt et~al.(2019)Warstadt, Singh, and
  Bowman}]{warstadt-etal-2019-neural}
Alex Warstadt, Amanpreet Singh, and Samuel~R. Bowman. 2019.
\newblock \href {https://doi.org/10.1162/tacl_a_00290} {Neural network
  acceptability judgments}.
\newblock \emph{Transactions of the Association for Computational Linguistics},
  7:625--641.

\bibitem[{Wolf et~al.(2019)Wolf, Debut, Sanh, Chaumond, Delangue, Moi, Cistac,
  Rault, Louf, Funtowicz, and Brew}]{Wolf2019HuggingFacesTS}
Thomas Wolf, Lysandre Debut, Victor Sanh, Julien Chaumond, Clement Delangue,
  Anthony Moi, Pierric Cistac, Tim Rault, R{\'e}mi Louf, Morgan Funtowicz, and
  Jamie Brew. 2019.
\newblock Huggingface's transformers: State-of-the-art natural language
  processing.
\newblock \emph{ArXiv}, abs/1910.03771.

\bibitem[{Zhao et~al.(2019)Zhao, Wang, Yatskar, Cotterell, Ordonez, and
  Chang}]{zhao-etal-2019-gender}
Jieyu Zhao, Tianlu Wang, Mark Yatskar, Ryan Cotterell, Vicente Ordonez, and
  Kai-Wei Chang. 2019.
\newblock \href {https://doi.org/10.18653/v1/N19-1064} {Gender bias in
  contextualized word embeddings}.
\newblock In \emph{Proceedings of the 2019 Conference of the North {A}merican
  Chapter of the Association for Computational Linguistics: Human Language
  Technologies, Volume 1 (Long and Short Papers)}, pages 629--634, Minneapolis,
  Minnesota. Association for Computational Linguistics.

\bibitem[{Zhao et~al.(2018)Zhao, Zhou, Li, Wang, and
  Chang}]{zhao-etal-2018-learning}
Jieyu Zhao, Yichao Zhou, Zeyu Li, Wei Wang, and Kai-Wei Chang. 2018.
\newblock \href {https://doi.org/10.18653/v1/D18-1521} {Learning gender-neutral
  word embeddings}.
\newblock In \emph{Proceedings of the 2018 Conference on Empirical Methods in
  Natural Language Processing}, pages 4847--4853, Brussels, Belgium.
  Association for Computational Linguistics.

\end{thebibliography}
\bibliographystyle{acl_natbib}

\appendix
\section*{Appendix}
\label{sec:appendix}

\subsection*{Datasets}

\paragraph{Sentence Encoder Association Test} 
(SEAT, \citet{may-etal-2019-measuring}) is based on Word Embedding Association Test (WEAT, \citet{Caliskan2017SemanticsDA}). Given two sets of \textit{attributes} and two sets of \textit{targets} words, WEAT measures differential cosine similarity between their embeddings. The two attribute sets can be male- and female-focused, where the targets can contain stereotypical associations, such as science- and arts-related vocabulary. SEAT extends the idea by embedding the vocabulary into sentences and taking their embedding representation (\texttt{[CLS]} classification token in case of transformer-based models). SEAT  measures bias only in the embedding space. That is, a model with a low SEAT score may still expose bias, as understood and perceived by humans. We employ SEAT6, -7, and -8 provided by \citet{may-etal-2019-measuring}.

\paragraph{StereoSet Stereotype Score}
(SS, \citet{nadeem-etal-2021-stereoset}) measures bias among four dimensions: gender, religion, occupation, and race. Technically, StereoSet is a dataset where each entry from four categories consists of a context and three options: stereotype, anti-stereotype and unrelated. 
On the top, StereoSet defines two tasks: \textit{intrasentence} and \textit{intersentence}. The objective of the former is to fill a gap with one of the options. The latter aims to choose a sentence that best follows the context. The SS score is a mean of scores on intra- and inter-sentence tasks.
Bias in StereoSet is measured as a “percentage of examples in which a model prefers a~stereotypical association [option] over an anti-stereotypical association” \cite{nadeem-etal-2021-stereoset}. An ideal bias-free model would have the bias score (\emph{stereotype score}, \emph{SS}) of 50. As opposed to SEAT, StereoSet SS models bias close to its human perception, as a preference of one thing over another.
We use the \textbf{gender} subset, as provided by \citet{nadeem-etal-2021-stereoset}.

\paragraph{General Language Understanding Evaluation} (GLUE, \citet{wang-etal-2018-glue})
is a popular benchmark to evaluate language model performance. It is a~suite of nine different tasks from domains such as sentiment analysis, paraphrasing, natural language inference, question answering, or sentence similarity. The GLUE score is an average of scores of all nine tasks.
To evaluate GLUE, we make use of the \texttt{run\_glue.py} script shipped by the Hugging Face library \cite{Wolf2019HuggingFacesTS}.

\paragraph{Gender Debiasing}
The debiasing algorithm introduced in $\S$\ref{sec:debiasing} requires some vocabulary lists.
We follow \citet{kaneko-bollegala-2021-debiasing}'s setup, that is we use lists of female and male attributes provided by \citet{zhao-etal-2018-learning}, and a list of stereotyped targets provided by \citet{kaneko-bollegala-2019-gender}.

\subsection*{Hyperparameters and Implementation}
For all experiments, we use the pre-trained \texttt{bert-base-uncased} \cite{devlin-etal-2019-bert} model from the open-source Hugging Face Transformers library (\citet{Wolf2019HuggingFacesTS}, ver. 4.12; \textit{Apache 2.0} license). 
We use $16$-bit floating-point mixed-precision training~\cite{Micikevicius2018MixedPT} as it halves training time and does not impact test performance.
To disentangle engineering from research, we use PyTorch Lightning framework (ver. 1.4.2;  \textit{Apache 2.0} license).
Model fine-pruning takes around 3h on a single A100 GPU.
All experiments can be reproduced with a random seed set to~$42$.

Usage of all libraries we used is consistent with their intended use.

\paragraph{Debiasing}
We provide an original implementation of the debiasing algorithm.
We use the same set of hyperparameters as \citet{kaneko-bollegala-2021-debiasing}, with an exception of a batch size of 128. We run debiasing (with no pruning - see \ref{tab:interm-debs}) for five epochs.

\paragraph{Pruning}
As for the pruning, we follow \citealt{lagunas-etal-2021-block}'s \textit{sigmoid-threshold} setting without the teacher network.
The threshold $\tau$ increases linearly from $0$ to $0.1$ over all training steps.
We fine-prune the BERT model with the debiasing objective for 100 epochs using a patched
\texttt{nn\_pruning}\footnote{\url{https://github.com/huggingface/nn_pruning/}} API (ver 0.1.2; \textit{Apache 2.0} license). See \texttt{README.md} in the attached code for instructions.

\subsection*{On Attention Head Pruning}
We cannot prune every matrix of the attention head if we want to prune the entire head.
To see why, let us recap the self-attention mechanism popularized by \citet{Vaswani2017AttentionIA}.

Denote an input sequence as $X\in\mathbb{R}^{N\times d}$, where $N$ is the sequence length and $d$ is a hidden size.
The first step of the self-attention is to obtain three matrices: $Q, K, V\in\mathbb{R}^{N\times d}$: \textit{queries}, \textit{keys}, and \textit{values}: $Q = XW^Q,   K = XW^K,   V = XW^V$,
where $W^Q, W^K, W^V\in\mathbb{R}^{d\times d}$ are learnable matrices.
The self-attention is defined as follows:
\begin{align*} 
    \text{SelfAtt}(Q, K, V) = 
         \text{softmax}
              \big(
              \frac{QK^T}{\sqrt{d}}
              \big)
              V.
\end{align*}

Now, suppose that the queries $W^Q$ or keys $W^K$ are pruned. Then the softmax would not cancel out the attention, but it would yield a uniform distribution over \textit{values} $W^V$.
Only by pruning values  $W^V$, we are able to make the attention output equal zero.

\section*{Supplementary Data}
Table \ref{tab:all-no-freeze} in this section show a breakdown of the GLUE scores, originally presented in \ref{tab:no-freeze}.
\begin{table}[h!]
\centering
\setlength{\tabcolsep}{2pt}
\resizebox{\linewidth}{!}{
\begin{tabular}{rrrrrrrrrrrr}
layer &  mode   & \ftn COLA & \ftn SST2 &\ftn MRPC &\ftn STSB &\ftn QQP  &\ftn MNLI &\ftn QNLI &\ftn RTE  &\ftn WNLI & GLUE \\[1pt] \hline \\[-10pt] 
all  & token    & \textbf{42.0} & 90.8 & 79.5 & 85.6 & 88.3 & \textbf{82.8} & \textbf{89.5} & \textbf{58.5} & 49.3 & 74.0 \\
     & sentence & 33.3 & 90.7 & 78.8 & 84.4 & 88.3 & 82.4 & 88.9 & 48.7 & 47.9 & 71.5 \\
last & token    & 41.3 & \textbf{91.1} & 80.5 & \textbf{85.7} & 88.5 & \textbf{82.8} & 89.3 & \textbf{58.5} & \textbf{52.1} & \textbf{74.4} \\
     & sentence & 40.2 & 90.6 & \textbf{80.7} & 85.2 & \textbf{88.5} & 81.9 & 88.2 & 49.8 & 45.1 & 72.2 \\
\end{tabular}
}
\caption{\label{tab:all-no-freeze}
Breakdown of the GLUE scores when fine-pruning BERT on the debiasing objective
with block size $32\times 32$ and letting the model's weights change freely.
}
\end{table}

\bigskip 

\section*{Bias Statement}
We follow \citet{kaneko-bollegala-2021-debiasing} and
define bias as stereotypical associations between male and female entities in pre-trained contextualized word representations. These representations when used for downstream applications, if not debiased, can further amplify gender inequalities \cite{komisyonu2020union}.
In our work, we focus on identifying layers of a language model that contribute to the biased associations.
We show that debiasing these layers can significantly reduce bias as measured in the embedding space (\textit{Sentence Encoder Association Test}, \citet{may-etal-2019-measuring}) and as perceived by humans, that is, as a preference of one thing over another (\textit{StereoSet Stereotype Score}, \citet{may-etal-2019-measuring}).
We limit our work solely to binary gender bias in the English language.

\end{document}